\documentclass[sigconf]{aamas}  

\usepackage{booktabs}
\usepackage{multirow}
\usepackage{subfig}
\usepackage{bm}
\usepackage{tabularx}                             
\usepackage{algorithm}
\usepackage{algorithmicx}
\usepackage{amssymb}
\usepackage[noend]{algpseudocode}
\usepackage{multicol}
\usepackage{amsmath}
\newcolumntype{L}[1]{>{\raggedright\arraybackslash}m{#1}}
\newcolumntype{P}[1]{>{\centering\arraybackslash}p{#1}}

\acmDOI{}  
\acmISBN{}  
\acmYear{2019}  
\copyrightyear{2019}  
\acmPrice{}  

\setcopyright{none}
\renewcommand\footnotetextcopyrightpermission[1]{}
\begin{document}

\title{Distributed Policy Iteration for Scalable Approximation of Cooperative Multi-Agent Policies}  



\author{Thomy Phan}
\affiliation{%
  \institution{LMU Munich}
}
\email{thomy.phan@ifi.lmu.de}

\author{Kyrill Schmid}
\affiliation{%
  \institution{LMU Munich}
  }
\email{kyrill.schmid@ifi.lmu.de}

\author{Lenz Belzner}
\affiliation{%
  \institution{MaibornWolff}
  }
\email{lenz.belzner@maibornwolff.de}

\author{Thomas Gabor}
\affiliation{%
  \institution{LMU Munich}
  }
\email{thomas.gabor@ifi.lmu.de}

\author{Sebastian Feld}
\affiliation{%
  \institution{LMU Munich}
  }
\email{sebastian.feld@ifi.lmu.de}

\author{Claudia Linnhoff-Popien}
\affiliation{%
  \institution{LMU Munich}
  }
\email{linnhoff@ifi.lmu.de}

\begin{abstract}  
Decision making in multi-agent systems (MAS) is a great challenge due to enormous state and joint action spaces as well as uncertainty, making centralized control generally infeasible. Decentralized control offers better scalability and robustness but requires mechanisms to coordinate on joint tasks and to avoid conflicts.
Common approaches to learn decentralized policies for cooperative MAS suffer from non-stationarity and lacking credit assignment, which can lead to unstable and uncoordinated behavior in complex environments.
In this paper, we propose \emph{Strong Emergent Policy approximation (STEP)}, a scalable approach to learn strong decentralized policies for cooperative MAS with a distributed variant of policy iteration. For that, we use function approximation to learn from action recommendations of a decentralized multi-agent planning algorithm. STEP combines decentralized multi-agent planning with centralized learning, only requiring a generative model for distributed black box optimization.
We experimentally evaluate STEP in two challenging and stochastic domains with large state and joint action spaces and show that STEP is able to learn stronger policies than standard multi-agent reinforcement learning algorithms, when combining multi-agent open-loop planning with centralized function approximation. The learned policies can be reintegrated into the multi-agent planning process to further improve performance.
\end{abstract}

%


\maketitle


\section{Introduction}\label{sec:introduction}

Cooperative multi-agent systems (MAS) are popular in artificial intelligence research and have many potential real-world applications like autonomous vehicles, sensor networks, and robot teams \cite{claes2015effective,claes2017decentralised,foerster2017counterfactual}. However, decision making in MAS is extremely challenging due to intractable state and joint action spaces as well as stochastic dynamics and uncertainty w.r.t. other agents' behavior.

Centralized control does not scale well in large MAS due to the \emph{curse of dimensionality}, where state and joint action spaces grow exponentially with the number of agents \cite{boutilier1996planning,amato2015scalable,claes2015effective,claes2017decentralised,gupta2017cooperative,foerster2017counterfactual}. Therefore, decentralized control is recommended, where each agent decides its individual actions under consideration of other agents, providing better scalability and robustness \cite{claes2015effective,claes2017decentralised,gupta2017cooperative,foerster2017counterfactual}. Decentralized approaches to decision making in MAS typically require a \emph{coordination mechanism} to solve joint tasks and to avoid conflicts \cite{boutilier1996planning}.

Learning decentralized policies with \emph{multi-agent reinforcement learning (MARL)} in cooperative MAS faces two major challenges: One challenge is \emph{non-stationarity}, where all agents adapt their behavior concurrently which can lead to unstable and uncoordinated policies \cite{laurent2011world,devlin2011theoretical,panait2006lenient,matignon2007hysteretic,foerster2018learning}. Another challenge is \emph{multi-agent credit assignment}, where the joint action of all agents leads to a single global reward which makes the deduction of the individual contribution of each agent difficult for adequate adaptation \cite{chang2004all,wolpert2002optimal,sunehag2017value,foerster2017counterfactual}.

Many approaches to solve these problems use reward or value decomposition to provide individual objectives \cite{chang2004all,gupta2017cooperative,sunehag2017value} or use reward shaping to obtain objectives which are easier to optimize \cite{devlin2014potential,wolpert2002optimal}. However, these approaches are generally not sufficient or infeasible due to complex emergent dependencies within large MAS which are hard to learn and specify \cite{foerster2017counterfactual}.

Recent approaches to learn strong policies are based on \emph{policy iteration} and combine planning with deep reinforcement learning, where a neural network is used to imitate the action recommendations of a tree search algorithm. In return, the neural network provides an action selection prior for the tree search \cite{anthony2017thinking,silver2017mastering}. This iterative procedure, called \emph{Expert Iteration (ExIt)}, gradually improves both the performance of the tree search and the neural network \cite{anthony2017thinking}. ExIt has been successfully applied to zero-sum games, where a single agent improves itself by self-play. However, ExIt cannot be directly applied to large cooperative MAS, since using a centralized tree search is practically infeasible for such problems \cite{claes2015effective,claes2017decentralised}.

In this paper, we propose \emph{Strong Emergent Policy approximation (STEP)}, a scalable approach to learn strong decentralized policies for cooperative MAS with a distributed variant of policy iteration. For that, we use function approximation to learn from action recommendations of a decentralized multi-agent planner. STEP combines decentralized multi-agent planning with centralized learning, where each agent is able to explicitly reason about emergent dependencies to make coordinated decisions. Our approach only requires a generative model for distributed black box optimization.

We experimentally evaluate STEP in two challenging and stochastic domains with large state and joint action spaces and show that STEP is able to learn stronger policies than standard MARL algorithms, when combining multi-agent open-loop planning with centralized function approximation. The policies can be reintegrated into the planning process to further improve performance.

The rest of the paper is organized as follows. Some background about decision making is provided in Section \ref{sec:background}. Section \ref{sec:related_work} discusses related work. STEP is described in Section \ref{sec:step}. Experimental results are presented and discussed in Section \ref{sec:experiments}. Section \ref{sec:conclusion} concludes and outlines a possible direction for future work.

\section{Background}\label{sec:background}
\subsection{Multi-Agent Markov Decision Processes}

\subsubsection{MDP}
A \emph{Markov Decision Process (MDP)} is defined by a tuple $M = \langle\mathcal{S},\mathcal{A},\mathcal{P},\mathcal{R}\rangle$, where $\mathcal{S}$ is a (finite) set of states, $\mathcal{A}$ is the (finite) set of actions, $\mathcal{P}(s_{t+1}|s_{t}, a_{t})$ is the transition probability function, and $\mathcal{R}(s_{t}, a_{t},s_{t+1}) \in \mathbb{R}$ is the reward function \cite{puterman2014markov}. We always assume that $s_{t}, s_{t+1} \in \mathcal{S}$, $a_{t} \in \mathcal{A}$, and $r_{t} = \mathcal{R}(s_{t}, a_{t},s_{t+1})$, where $s_{t+1}$ is reached after executing $a_{t}$ in $s_{t}$ at time step $t$.

The goal is to find a \emph{policy} $\pi : \mathcal{S} \rightarrow \mathcal{A}$ which maximizes the (discounted) return $G_{t}$ at state $s_{t}$ for a horizon $h$:
\begin{equation}\label{eq:return}
G_{t} = \sum_{k=0}^{h-1} \gamma^{k} \mathcal{R}(s_{t+k}, a_{t+k}, s_{t+k+1})
\end{equation}
where $\gamma \in [0,1)$ is the discount factor. Alternatively, a policy may be stochastic such that $\pi(a_{t}|s_{t}) \in [0,1]$ with $\sum_{a_{t} \in \mathcal{A}}^{}\pi(a_{t}|s_{t}) = 1$.

A policy $\pi$ can be evaluated with a \emph{state value function} $V^{\pi}(s_{t}) = \mathbb{E}_{\pi}[G_{t}|s_{t}] = \textit{max}_{a_{t} \in \mathcal{A}}(Q^{\pi}(s_{t},a_{t}))$, which is defined by the expected return at $s_{t}$. $Q^{\pi}(s_{t},a_{t}) = \mathbb{E}_{\pi}[G_{t}|s_{t},a_{t}]$ is the \emph{action value function} of $\pi$ defining the expected return when executing $a_{t}$ in $s_{t}$.

$\pi$ is optimal, if it is \emph{stronger} than all other policies $\pi'$ such that $V^{\pi}(s_{t}) \geq V^{\pi'}(s_{t})$ for all $s_{t} \in S$. We denote the optimal policy by $\pi^{*}$ and the optimal value function by $V^{\pi^{*}} = V^{*}$ or $Q^{\pi^{*}} = Q^{*}$ resp.

\subsubsection{MMDP}
An MDP can be extended to a \emph{multi-agent MDP (MMDP)} with a (finite) set of agents $\mathcal{D} = \{1,...,N\}$. $\mathcal{A} = \mathcal{A}_{1} \times ... \times \mathcal{A}_{N}$ is the (finite) set of \emph{joint actions}. The goal is to find a \emph{joint policy} $\pi = \langle\pi_{1},...,\pi_{N}\rangle$ which maximizes the return $G_{t}$ of Eq. \ref{eq:return}. $\pi_{i}$ is the \emph{local policy} of agent $i \in \mathcal{D}$. Similarly to MDPs, a value function $V^{\pi}$ can be used to evaluate the joint policy $\pi$.

MMDPs can be used to model fully observable problems for cooperative MAS, where all agents share a common goal \cite{boutilier1996planning,claes2015effective,claes2017decentralised}. In this paper, we focus on \emph{homogeneous} MAS with $\mathcal{A}_{1} = ... = \mathcal{A}_{N}$ and $\pi^{*} = \langle\pi_{1}^{*},...,\pi_{N}^{*}\rangle$, where $\pi_{i}^{*} = \pi_{j}^{*}$, even if $i \neq j$ \cite{panait2005cooperative,varakantham2012decision,zinkevich2001symmetry} \footnote{We assume that each agent is uniquely identified by its identifier $i$ and individual state $s_{t}^{i}$ to ensure that there exists an optimal local policy $\pi_{i}^{*}$ for all agents \cite{shoham1995social,boutilier1996planning,zinkevich2001symmetry}.}.

\subsection{Planning}
\emph{Planning} searches for an (near-)optimal policy, given a model $\hat{M}$ of the environment $M$. $\hat{M}$ provides an approximation for $\mathcal{P}$ and $\mathcal{R}$ of the underlying MDP or MMDP \cite{boutilier1996planning}. \emph{Global planning} searches the whole state space $\mathcal{S}$ to find $\pi^{*}$. \emph{Policy iteration} is a global planning approach which computes $\pi^{*}$ with alternating \emph{policy evaluation}, where $V^{\pi^{n}}$ is computed for the current policy $\pi^{n}$ and \emph{policy improvement}, where a \emph{stronger policy} $\pi^{n+1}$ is generated by selecting actions that maximize $V^{\pi^{n}}$ for each state $s_{t} \in \mathcal{S}$ \cite{beranek1961ronald,boutilier1996planning}.
\emph{Local planning} only regards the current state $s_{t}$ and possible future states to find a policy $\pi_{t}$ with closed- or open-loop search \cite{weinstein2013open}.
\emph{Monte Carlo planning} uses a \emph{generative model} $\hat{M}$ as black box simulator without reasoning about explicit probability distributions \cite{kocsis2006bandit,weinstein2013open,lecarpentier2018open}.

\emph{Closed-loop planning} conditions the action selection on the history of previous states and actions. \emph{Monte Carlo Tree Search (MCTS)} is a popular closed-loop algorithm, which incrementally constructs a search tree to estimate $Q^{*}$ \cite{kocsis2006bandit,silver2017mastering}. It traverses the tree by selecting nodes $s_{t} \in \mathcal{S}$ with a policy $\pi_{\textit{tree}}$ until a leaf node $s_{h-1}$ is reached. $\pi_{\textit{tree}}$ is commonly implemented with the UCB1 selection strategy which maximizes $\textit{UCB1} = \overline{Q(s_{t},a_{t})} + c \sqrt{2\textit{log}(n_{t}) / n_{a_{t}}}$,
where $\overline{Q(s_{t},a_{t})}$ is the average return, $n_{t}$ is the visit count of $s_{t}$, $n_{a_{t}}$ is the selection count of $a_{t}$, and $c$ is an exploration constant \cite{auer2002finite,kocsis2006bandit}.
The node $s_{h-1}$ is expanded by a new child node $s_{h}$, whose value $\hat{V}(s_{h})$ is estimated with a rollout or a value function \cite{silver2017mastering}. The observed rewards are recursively accumulated to returns $G_{t}$ (Eq. \ref{eq:return}) to update the value estimate $\overline{Q(s_{t},a_{t})}$ of each state-action pair in the search path. MCTS is an \emph{anytime algorithm}, which returns an action recommendation for the root state $s_{0}$ 
according to the highest action value
after a \emph{computation budget} $n_{b}$ has run out.

\begin{figure}[!ht]
     \subfloat[closed-loop tree\label{fig:closed_loop_planning}]{%
       \includegraphics[width=0.25\textwidth]{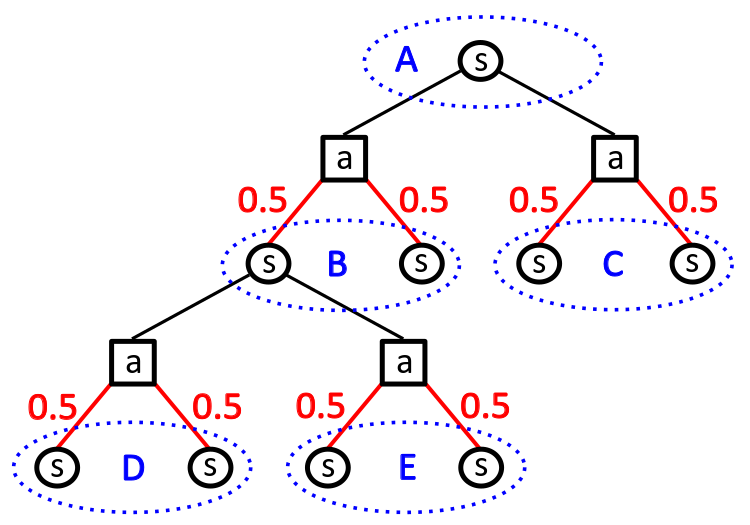}
     }
     \hfill
     \subfloat[open-loop tree\label{fig:open_loop_planning}]{%
       \includegraphics[width=0.16\textwidth]{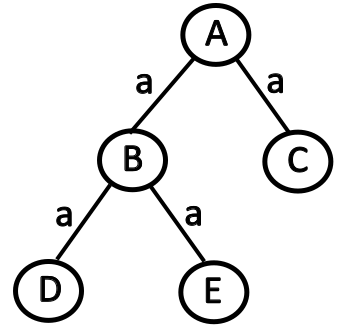}
     }
     \caption{Illustration of closed- and open-loop trees for planning. (a) Closed-loop tree with states (circular nodes) and actions (rectangular nodes). Red links correspond to stochastic state transitions with a probability of 0.5. (b) Open-loop tree with links as actions and nodes as state distributions according to the blue dotted ellipses in Fig. \ref{fig:closed_loop_planning}.}
     \label{fig:closed_vs_open_loop_planning}
\end{figure}

In stochastic domains, closed-loop planning needs to store each state $s_{t+1}$ encountered when executing $a_{t}$ in $s_{t}$. This may lead to large search trees with high branching factors, if $\mathcal{S}$ and $\mathcal{A}$ are very large. \emph{Open-loop planning} conditions the action selection only on previous actions and summarized statistics about predecessor states, thus reducing the search space \cite{weinstein2013open,perez2015open}. An example is shown in Fig. \ref{fig:closed_vs_open_loop_planning}. A closed-loop tree for a domain with $\mathcal{P}(s_{t+1}|s_{t},a_{t}) = 0.5$ is shown in Fig. \ref{fig:closed_loop_planning}. Fig. \ref{fig:open_loop_planning} shows the corresponding open-loop tree which summarizes the state nodes of Fig. \ref{fig:closed_loop_planning} within the blue dotted ellipses into state distribution nodes.
\emph{Open-Loop UCB applied to Trees (OLUCT)} is an open-loop variant of MCTS with UCB1 \cite{lecarpentier2018open}.

\subsection{Reinforcement Learning}\label{subsec:reinforcement_learning}
\emph{Reinforcement Learning (RL)} searches for an (near-)optimal policy in an environment $M$ without knowing the effect of executing $a_{t} \in \mathcal{A}$ in $s_{t} \in \mathcal{S}$ \cite{boutilier1996planning,sutton1998introduction}. RL agents typically obtain experience samples $E = \{e_{1},...,e_{t}\}$ with $e_{t} = \langle s_{t}, a_{t}, s_{t+1},r_{t}\rangle$ by interacting with the environment.
\emph{Model-based} RL methods learn a model $\hat{M} \approx M$ by approximating $\mathcal{P}$ and $\mathcal{R}$ with $E$ \cite{boutilier1996planning,sutton1998introduction}. $\hat{M}$ can be used for planning to find a policy. 
Alternatively, $\pi^{*}$, $V^{*}$, and/or $Q^{*}$ can be approximated directly with $E$ by using \emph{model-free RL} \cite{watkins1992q,sutton1988learning,sutton2000policy,konda2000actor}.

\subsection{Decision Making in MMDPs}
An MMDP can be formulated as a joint action MDP and solved with single-agent planning or RL by directly searching for joint policies \cite{boutilier1996planning}. However, this does not scale well due to the curse of dimensionality, where the state and joint action spaces grow exponentially with the number of agents \cite{boutilier1996planning,amato2015scalable,claes2015effective,claes2017decentralised,gupta2017cooperative,foerster2017counterfactual}.

Alternatively, local policies can be searched with decentralized planning or RL, where each agent plans or learns independently of each other \cite{tan1993multi,claes2017decentralised}. Decentralized approaches typically require a \emph{coordination mechanism} to solve joint tasks and to avoid conflicts \cite{boutilier1996planning}. Common mechanisms are communication to exchange private information \cite{tan1993multi,phan2018evade}, synchronization to reach a consensus \cite{emery2004approximate,omidshafiei2017deep}, or prediction of other agents' behavior with policy models \cite{claes2015effective,claes2017decentralised}.

\section{Related Work}\label{sec:related_work}

\paragraph{Policy Iteration with Deep Learning and Tree Search}
Recently, approaches to learning strong policies from MCTS recommendations with deep learning have been successfully applied to single-agent domains and zero-sum games  \cite{guo2014deep,anthony2017thinking,silver2017mastering,jiang2018feedback}. In zero-sum games, a single agent is trained via self-play, gradually improving itself when playing against an increasingly stronger opponent. This corresponds to the policy iteration scheme, where self-play evaluates the current policy and MCTS improves the policy by recommending stronger actions based on the evaluation.

Our approach addresses cooperative problems with \emph{multiple} agents, where a local policy has to be learned for each agent, which maximizes the \emph{common return}. We use \emph{decentralized multi-agent planning} to recommend actions for local policy approximation, since centralized planning would be infeasible for large MAS \cite{boutilier1996planning,claes2017decentralised}. The local policy approximation also serves as \emph{coordination mechanism} to predict other agents' behavior during planning \cite{varakantham2012decision,claes2015effective,claes2017decentralised}.

\paragraph{Policy Iteration for Cooperative MAS}
Previous work on policy iteration in MAS has focused on centralized offline planning, where an (near-)optimal joint policy is searched by exhaustively evaluating and updating all local policy candidates for each agent with an explicit model of the MAS \cite{hansen2004dynamic,bernstein2005bounded,szer2006point,bernstein2009policy}. Dominated policy candidates can be discarded by using heuristics to reduce computation \cite{seuken2007memory,wu2012sample}. However, these approaches do not scale well for complex domains due to the curse of dimensionality of the (joint) policy space \cite{oliehoek2016concise}.

Our approach is more scalable, since we use \emph{decentralized multi-agent planning} for training, which can be performed online, and a single \emph{function approximation} to learn a local policy for each agent. Our approach only requires a \emph{generative model} for black box optimization but no explicit probability distributions of the MAS.

\paragraph{Multi-Agent Reinforcement Learning (MARL)}
MARL is a widely studied field \cite{tan1993multi,bucsoniu2010multi} and has been often combined with deep learning \cite{foerster2016learning,tampuu2017multiagent}.
A scalable and popular way to cooperative MARL is to let each agent learn its local policy independently of others \cite{tan1993multi,tampuu2017multiagent,leibo2017multi}, but non-stationarity and the lack of credit assignment can lead to uncoordinated behavior \cite{devlin2011theoretical,laurent2011world,sunehag2017value,foerster2017counterfactual}. Non-stationarity can be addressed with stabilized or synchronized experience replay \cite{omidshafiei2017deep,foerster2017stabilising}, or with opponent modeling \cite{he2016opponent,rabinowitz18theory,hong2018deep,zhang2010multi,foerster2018learning}. Approaches to credit assignment provide local rewards for each agent \cite{gupta2017cooperative,lin2018efficient}, learn a filtering of local rewards from a global reward \cite{chang2004all,sunehag2017value}, or use reward shaping \cite{devlin2014potential,wolpert2002optimal,foerster2017counterfactual}. In many cases, learning is centralized, where all agents share experience or parameters to accelerate learning of coordinated local policies \cite{tan1993multi,foerster2016learning,gupta2017cooperative,foerster2017counterfactual}. While training might be centralized, the execution of the policies is decentralized \cite{foerster2016learning,gupta2017cooperative}.

Our approach learns a local policy from \emph{action recommendations} of a \emph{decentralized multi-agent planner}. A generative model is used for distributed black box optimization, where each agent \emph{explicitly reasons} about the global effect of its \emph{individual actions} without additional domain knowledge or reward decomposition.

\section{STEP}\label{sec:step}
We now describe \emph{Strong Emergent Policy approximation (STEP)} for learning strong decentralized policies by using a distributed variant of policy iteration. STEP defines a framework to combine decentralized multi-agent planning with centralized learning.

\subsection{Scalable Policy Iteration for MAS}\label{subsec:scalable_policy_iteration}

Similarly to MDPs, policy iteration for MMDPs consists of an alternating \emph{evaluation} and \emph{improvement} step \cite{beranek1961ronald,boutilier1996planning}. Given a joint policy $\pi^{n} = \langle \pi_{1}^{n}, ..., \pi_{N}^{n} \rangle$, the global value function $V^{\pi^{n}}$ can be computed to evaluate $\pi^{n}$. By selecting joint actions $a_{t} = \langle a_{t,1}, ..., a_{t,N} \rangle \in \mathcal{A}$ which maximize $V^{\pi^{n}}$ for each state $s_{t} \in \mathcal{S}$, we obtain an improved joint policy $\pi^{n+1}(s_{t}) = a_{t}$, which is stronger than $\pi^{n}$. In MMDPs, the state and joint action spaces are typically too large to exactly compute $V^{\pi^{n}}$ and $\pi^{n+1}$ \cite{boutilier1996planning,claes2017decentralised}. Thus, we use function approximation to compute $\hat{V} \approx V^{\pi^{n}}$ and $\hat{\pi} = \langle \hat{\pi_{1}}, ..., \hat{\pi_{N}} \rangle \approx \pi^{n+1}$ \cite{anthony2017thinking,silver2017mastering}. 

For \emph{scalable policy evaluation}, we use temporal difference (TD) learning to train $\hat{V}$ with real experience and ensure generalization to avoid computing $V^{\pi^{n}}(s_{t})$ for each state $s_{t} \in \mathcal{S}$ explicitly \cite{mnih2015human}.

For \emph{scalable policy improvement}, we use decentralized multi-agent planning, where each agent $i$ explicitly reasons about the global effect of its individual actions $a_{t,i} \in \mathcal{A}_{i}$ to maximize the common return $G_{t}$ instead of searching the whole joint action space $\mathcal{A}$. A function approximator $\hat{\pi}_{i}$ is used to learn a local policy from the action recommendations of each agent's individual planner.

The explicit reasoning mitigates the credit assignment problem, since each agent $i$ is incentivized to optimize its individual actions to maximize the common return based on the global value function $\hat{V}$ and the policy $\hat{\pi}_{j}$ of all other agents $j \neq i$. Since $\hat{\pi}_{i}$ is trained to imitate the individual planner of agent $i$, $\hat{\pi}_{i}$ can be used to predict future actions of agent $i$, similarly to opponent modeling to address non-stationarity, when optimizing local decisions. This can lead to coordinated actions to solve joint tasks and to avoid conflicts \cite{claes2017decentralised}.

Combining these elements leads to a \emph{distributed policy iteration} scheme for cooperative MAS, which only requires a generative model for distributed black box optimization.

\subsection{Decentralized Open-Loop UCT}\label{subsec:DOLUCT}

Decentralized closed-loop planning, with a worst-case branching factor of $b_{\textit{cl}} = |\mathcal{S}|\cdot|\mathcal{A}_{i}|$, quickly becomes infeasible when the problem is too large to provide sufficient computation budget. Thus, we focus on open-loop planning, which generally explores much smaller search spaces with a branching factor of $b_{\textit{ol}} = |\mathcal{A}_{i}| << b_{\textit{cl}}$ (Fig. \ref{fig:closed_vs_open_loop_planning}), and can be competitive to closed-loop planning, when computational resources are highly restricted \cite{weinstein2013open,perez2015open,lecarpentier2018open}.
We propose a decentralized variant of OLUCT from \cite{lecarpentier2018open}, which we call \emph{DOLUCT}.

At every time step $t$, all agents perform an independent DOLUCT search in parallel. A stochastic policy function $\hat{\pi}_{i}(a_{t,i}|s_{t}) \in [0,1]$ is used to simulate all other agents. To traverse a DOLUCT search tree, we propose a modified version of UCB1 similarly to \cite{silver2017mastering}:
\begin{equation}\label{eq:ucb1_STEP}
\textit{UCB1}_{\textit{DOLUCT}}^{\hat{\pi}_{i}}(\textit{Nd}_{t},s_{t},a_{t,i}) = \overline{Q(\textit{Nd}_{t},a_{t,i})} + \hat{\pi}_{i}(a_{t,i}|s_{t}) c \sqrt{\frac{2\textit{log}(n_{t,i})}{n_{a_{t,i}}}}
\end{equation}
where $\textit{Nd}_{t}$ is a node in the open-loop tree (Fig. \ref{fig:open_loop_planning}). Note that the local action probabilities $\hat{\pi}_{i}(a_{t,i}|s_{t})$ for the same node $\textit{Nd}_{t}$ can vary depending on the currently simulated state $s_{t}$, thus providing a \emph{closed-loop prior} for the action selection.

To avoid searching the full depth of the problem, we propose to use a value function $\hat{V}$ to evaluate states at leaf nodes \cite{silver2017mastering,phan2018evade}.

The complete formulation of DOLUCT is given in Algorithm \ref{algorithm:DOLUCT}, where $i$ identifies the current agent, $s_{t}$ is the state to plan on, $\hat{M}$ is the generative model, $N$ is the number of agents, $n_{b}$ is the computation budget, $\hat{V}$ is a value function, and $\hat{\pi}_{i}$ is a local policy.

\begin{algorithm}
\caption{Decentralized Open-Loop UCT (DOLUCT)}\label{algorithm:DOLUCT}
\begin{algorithmic}[1]
\Procedure{$\textit{DOLUCT}(i, s_{t}, \hat{M}, N, n_{b}, \hat{V}, \hat{\pi}_{i})$}{}
\State Initialize $\textit{Nd}_{t}$: $\overline{Q(\textit{Nd}_{t},a_{t,i})} = n_{t,i} = n_{a_{t,i}} = 0, \forall a_{t,i} \in \mathcal{A}_{i}$ 
\While{$n_{b} > 0$}
\State $\langle n_{b},G_{t}\rangle \leftarrow \textit{Simulate}(i, \textit{Nd}_{t}, s_{t}, \hat{M}, N, n_{b}, \hat{V}, \hat{\pi}_{i})$
\State $p(a_{t,i}|s_{t}) \leftarrow \frac{n_{a_{t,i}}}{n_{t,i}}, \forall a_{t,i} \in \mathcal{A}_{i}$
\EndWhile
\Return $\langle \textit{argmax}_{a_{t,i} \in \mathcal{A}_{i}}(\overline{Q(\textit{Nd}_{t},a_{t,i})}), p(a_{t,i}|s_{t}) \rangle$
\EndProcedure
\end{algorithmic}
\begin{algorithmic}[1]
\Procedure{$Simulate(i, {Nd}_{t}, s_{t}, \hat{M}, N, n_{b}, \hat{V}, \hat{\pi}_{i})$}{}
\If{$n_{b} \leq 0$}
\Return $\langle 0, \hat{V}(s_{t}) \rangle$
\EndIf
\If{$\textit{Nd}_{t}$ is a leaf node}
\State Expand $\textit{Nd}_{t}$
\State
\Return $\langle n_{b}, \hat{V}(s_{t}) \rangle$
\EndIf
\State $a_{t,i} \leftarrow \textit{argmax}_{a_{t,i} \in \mathcal{A}_{i}}(\textit{UCB1}_{\textit{DOLUCT}}^{\hat{\pi}_{i}}(\textit{Nd}_{t},s_{t},a_{t,i}))$
\State $a_{t,j} \sim \hat{\pi}_{i}(a_{t,j}|s_{t}), \forall_{j \in \{1,...,N\}} j \neq i$ \Comment{simulate other agents}
\State $a_{t} \leftarrow \langle a_{t,1}, ..., a_{t,N} \rangle$
\State $\langle s_{t+1}, r_{t} \rangle \sim \hat{M}(s_{t}, a_{t})$
\State $\langle n_{b}, n_{t,i}, n_{a_{t,i}} \rangle \leftarrow \langle n_{b} - 1, n_{t,i} + 1, n_{a_{t,i}} + 1 \rangle$
\State $\langle n_{b}, R_{t} \rangle \leftarrow \textit{Simulate}(i, \textit{Nd}_{t+1}, s_{t+1}, \hat{M}, N, n_{b}, \hat{V}, \hat{\pi}_{i})$
\State $G_{t} \leftarrow r_{t} + \gamma R_{t}$
\State $\overline{Q(\textit{Nd}_{t},a_{t,i})} \leftarrow ((n_{t,i} - 1) \overline{Q(\textit{Nd}_{t},a_{t,i})} + G_{t})/n_{t,i}$
\State
\Return $\langle n_{b}, G_{t} \rangle$
\EndProcedure
\end{algorithmic}
\end{algorithm}

\subsection{Strong Emergent Policy Approximation}\label{subsec:STEP_approach}
We intend to learn $\pi_{i}^{*}$ by imitating a decentralized multi-agent planning algorithm similarly to ExIt \cite{anthony2017thinking,silver2017mastering} for the single-agent case. The planner itself is improved with $\hat{\pi}_{i} \approx \pi_{i}^{*}$ as an action selection prior and as prediction of other agents' behavior for coordination, and $\hat{V} \approx V^{*}$ as leaf state evaluator. We assume an online setting with an alternating \emph{planning} and \emph{learning} step for each time step $t$.

In the planning step, a joint action $a_{t} = \langle a_{t,1}, ..., a_{t,N} \rangle$ is searched with decentralized multi-agent planning for the current state $s_{t}$. The planning algorithm can exploit the policy approximation $\hat{\pi}_{i}$ as a prior for action selection (e.g., Eq. \ref{eq:ucb1_STEP}) and as prediction of other agents' behavior for coordination. All agents execute $a_{t}$ and cause a state transition to $s_{t+1}$, while observing a global reward $r_{t}$. The transition $e_{t} = \langle s_{t}, a_{t}, s_{t+1}, r_{t}, p_{t} \rangle$ is stored as \emph{experience sample} in a \emph{central buffer} $E$, where $p_{t} = \langle p(a_{t,1}|s_{t}),..., p(a_{t,N}|s_{t}) \rangle$ contains the relative frequencies of the action selections $p(a_{t,i}|s_{t}) = \frac{n_{a_{t,i}}}{n_{t,i}}, \forall a_{t,i} \in \mathcal{A}_{i}$ of each agent's individual planner for state $s_{t}$.

In the learning step, a parametrized function approximator $f_{i,\theta} = \langle \hat{\pi}_{i}, \hat{V} \rangle$ with parameter vector $\theta$ is used to approximate $\pi_{i}^{*}$ and $V^{*}$ by minimizing the loss $L_{\textit{STEP}} =  \mathbb{E}_{i \in \mathcal{D}}[L_{\textit{STEP},i}]$ for all agents of all transitions $e_{t} \in E$ w.r.t. $\theta$:
\begin{equation}\label{eq:step_loss_function}
L_{\textit{STEP},i} = (y_{t} - \hat{V}(s_{t}))^{2} - p(a_{t,i}|s_{t})^{\top}\textit{log} (\hat{\pi}_{i}(a_{t,i}|s_{t}))
\end{equation}
where $y_{t} = r_{t} + \gamma \hat{V}(s_{t+1})$ is the TD target for $\hat{V}$ \cite{sutton1988learning}.
$p(a_{t,i}|s_{t})$ and $\hat{\pi}_{i}(a_{t,i}|s_{t})$ are $|\mathcal{A}_{i}|$-dimensional probability vectors. The first term is the squared TD error and the second term is the cross entropy between $p(a_{t,i}|s_{t})$ and $\hat{\pi}_{i}(a_{t,i}|s_{t})$. With $L_{\textit{STEP}}$, $f_{i,\theta}$ can be trained online because it can incrementally incorporate new experience $e_{t}$ \footnote{In practice, $\theta$ is optimized on random experience batches of constant size \cite{mnih2015human}.}. Thus, $f_{i,\theta}$ is able to adapt to changes at system runtime and does not require the problem to be episodic \cite{sutton1988learning,phan2018evade}.

The complete formulation of STEP is given in Algorithm \ref{algorithm:STEP}, where $\textit{DPlan}$ is a decentralized multi-agent planning algorithm, $\hat{M}$ is the generative model, $N$ is the number of agents, $n_{b}$ is the computation budget, and $f_{i,\theta}$ is the function approximator for $\pi_{i}^{*}$ and $V^{*}$.

\begin{algorithm}
\caption{Strong Emergent Policy Approximation (STEP)}\label{algorithm:STEP}
\begin{algorithmic}[1]
\Procedure{$\textit{STEP}(DPlan, \hat{M}, N, n_{b}, f_{i,\theta})$}{}
\State Initialize $\theta$ of $f_{i,\theta}$
\State Observe $s_{1}$
\For{$t = 1,T$} \Comment{$T$ is the training duration (can be infinite)}
\State $\langle \hat{\pi}_{i}, \hat{V} \rangle \leftarrow f_{i,\theta}$ 
\For{$i \in \{1, ..., N\}$} \Comment{decentralized planning}
\State	$\langle a_{t,i}, p(a_{t,i}|s_{t}) \rangle \leftarrow \textit{DPlan}(i, s_{t}, \hat{M}, N, n_{b}, \hat{V}, \hat{\pi}_{i})$
\EndFor
\State  $p_{t} \leftarrow \langle p(a_{t,1}|s_{t}),..., p(a_{t,N}|s_{t}) \rangle$
\State	Execute $a_{t} \leftarrow \langle a_{t,1}, ..., a_{t,N} \rangle$
\State  Observe global reward $r_{t}$ and new state $s_{t+1}$
\State	Store $e_{t} \leftarrow \langle s_{t}, a_{t}, s_{t+1}, r_{t}, p_{t}\rangle$ in $E$
\State	Refine $\theta$ to minimize $L_{\textit{STEP}}$ for all $e_{t} \in E$ \Comment{centralized learning}
\EndFor
\EndProcedure
\end{algorithmic}
\end{algorithm}

\subsection{STEP Architecture}
\begin{figure}
  \centering
    \includegraphics[width=0.35\textwidth]{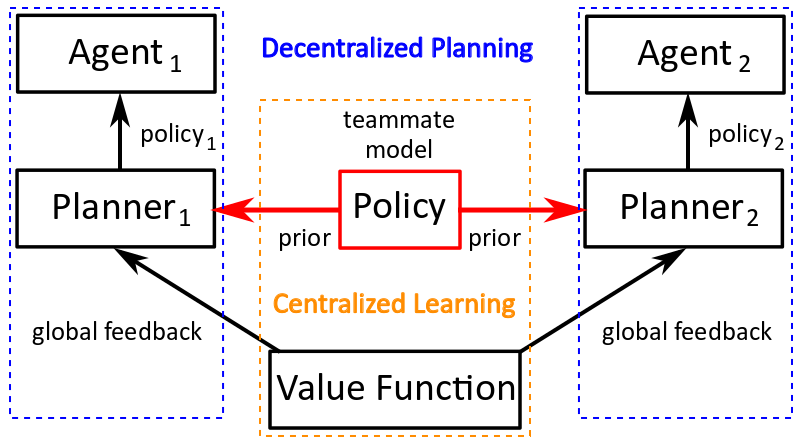}
    \caption{Architecture of STEP. The policy (red box) can be used as a prior for action selection and to predict other agents' behavior for coordination. The value function is used to evaluate leaf states for multi-agent planning \cite{phan2018evade}. }\label{fig:step_architecture}
\end{figure}

Fig. \ref{fig:step_architecture} shows the conceptual architecture for two agents. All agents are controlled by individual planners, which share the same local policy $\hat{\pi}_{i}$ and the global value function $\hat{V}$. $\hat{\pi}_{i}$ is used as an action selection prior and as prediction of other agents' behavior for coordination. $\hat{V}$ is used to evaluate leaf states during planning.

Although multi-agent planning is \emph{decentralized}, learning is \emph{centralized} to accelerate the approximation \cite{tan1993multi,foerster2016learning,gupta2017cooperative,foerster2017counterfactual}. The parameters are shared among all agents and updated in a centralized manner. Although we focus on homogeneous MAS, where all agents use the same local policy $\hat{\pi}_{i}$, our approach can be easily extended to heterogeneous MAS. Given a MAS with $K$ different agent types, $K$ different local policies need to be approximated (one for each agent type), which can still be done in a centralized fashion such that all agents have access to these $K$ policies during training.

\subsection{Bias Regulation}\label{subsec:bias_regulation}
Using $f_{i,\theta}$ in decentralized tree search algorithms (e.g., DOLUCT) induces a bias in estimating $\pi_{i}^{*}$ and $V^{*}$ with $\hat{\pi_{i}}$ and $\hat{V}$ resp., thus includes approximation errors in the planning step.

The action selection bias of $\hat{\pi_{i}}$ can be reduced by increasing the computation budget $n_{b}$. The more node $\textit{Nd}_{t}$ is visited, the smaller the exploration term multiplied with $c$ in Eq. \ref{eq:ucb1_STEP} becomes, which decreases the influence of $\hat{\pi_{i}}$. This causes the search to focus on nodes with higher expected return $\mathbb{E}[G_{t}|\textit{Nd}_{t}]$, thus the search tree is expanded into these directions. The increasing horizon $h$ discounts the value estimate $\hat{V}(s_{h})$ of newly added nodes $\textit{Nd}_{h}$ by a factor of $\gamma^{h}$, thus reducing the bias of $\hat{V}$ for frequently visited paths, if $\gamma < 1$.

\section{Experiments}\label{sec:experiments}
\subsection{Evaluation Environments}\label{subsec:environments}

\begin{figure}[!ht]
     \subfloat[Pursuit \& Evasion\label{fig:pursuit_evasion_map}]{%
       \includegraphics[width=0.14\textwidth]{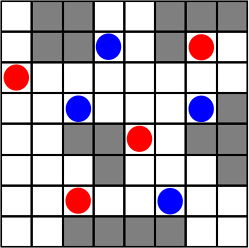}
     }
     \hfill
     \subfloat[Smart Factory (SF)\label{fig:machine_grid}]{%
       \includegraphics[width=0.1452\textwidth]{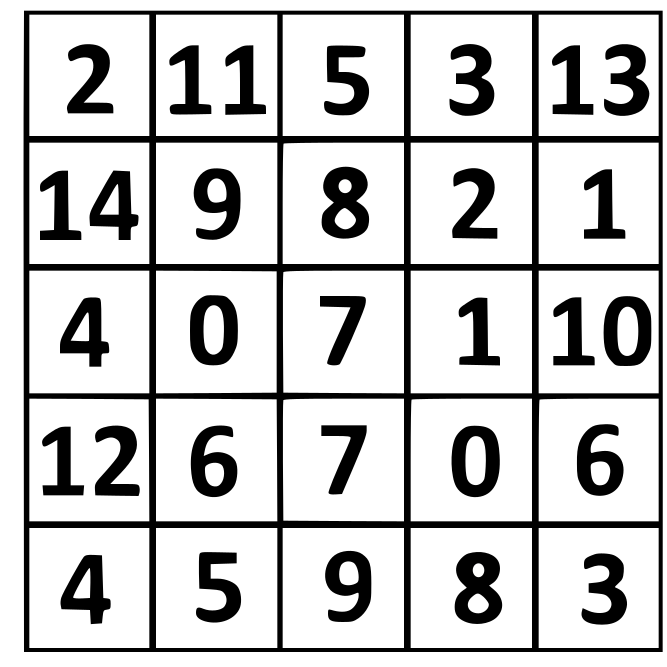}
     }
     \hfill
     \subfloat[An agent in SF \cite{phan2018evade}\label{fig:item_tasks_example}]{%
       \includegraphics[width=0.1452\textwidth]{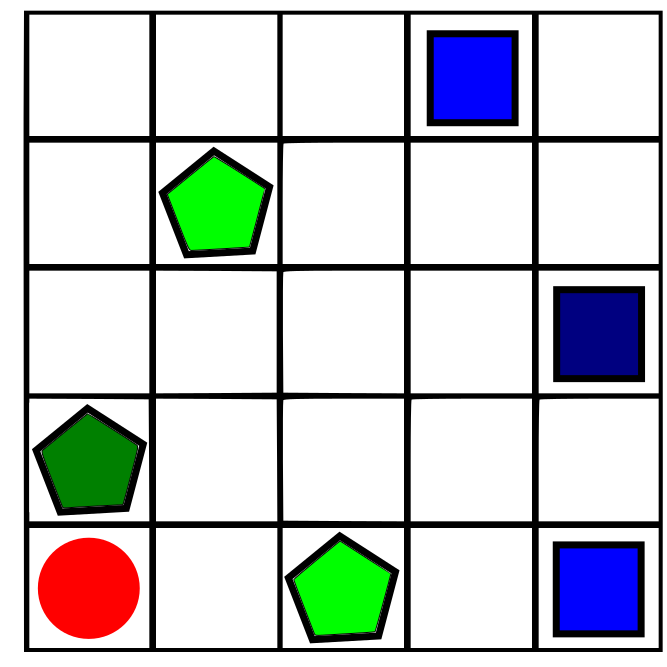}
     }
     \caption{(a) Pursuit \& Evasion ($N = 4$) with pursuers (red circles) and evaders (blue circles). (b) Machine grid of the Smart Factory (SF) with the numbers denoting the machine type. (c) An agent $i$ (red circle) with $tasks_{i} = [\{9, 12\},\{3, 10\}]$ in the SF of Fig. \ref{fig:machine_grid}. It should get to the green pentagonal machines first before going to the blue rectangular machines \cite{phan2018evade}.}
     \label{fig:evaluation_environments}
\end{figure}

\subsubsection{Pursuit \& Evasion (PE)}
PE is a well-known benchmark problem for MARL algorithms \cite{tan1993multi,vidal2002probabilistic,gupta2017cooperative}. We implemented this domain as $8 \times 8$ grid with $N$ \emph{pursuers} as learning agents, $N$ \emph{evaders} as randomly moving entities, and some obstacles as shown in Fig. \ref{fig:pursuit_evasion_map} for $N = 4$, where the pursuers must collaborate to capture all evaders. All pursuers and evaders have random initial positions and are able to move north, south, west, east, or do nothing. If two pursuers occupy the same cell as an evader, a global reward of $r_{t} = 1$ is obtained.
The reward $r_{t} = \sum_{i = 1}^{N} r_{t,i}$ can be decomposed into local rewards $r_{t,i}$, where each of the $\textit{N}_{\textit{hunter}}$ pursuers $i$, which occupied the cell of the captured evader, is rewarded with $r_{t,i} = \frac{1}{\textit{N}_{\textit{hunter}}}$.

\subsubsection{Smart Factory (SF)}
SF was introduced in \cite{phan2018evade}. It consists of a $5 \times 5$ grid of machines as shown in Fig. \ref{fig:machine_grid} and $N$ agents with each agent having one \emph{item}, a list of randomly assigned tasks $\textit{tasks}_{i}$ organized in \emph{buckets}, and a random initial position. An example from \cite{phan2018evade} is shown in Fig. \ref{fig:item_tasks_example} for an agent $i$ with $\textit{tasks}_{i} = [\{9, 12\},\{3, 10\}]$. It first needs to get processed at the machines having a machine type of 9 and 12 (green pentagons) before going to the machines with type 3 and 10 (blue rectangles). All agents are able to enqueue at their current machine, move north, south, west, east, or do nothing. At every time step, each machine processes one agent in its queue with a cost of 0.25 but does nothing with a probability of 0.1. If a task in the current bucket of the processed agent matches with the machine's type, the task is removed from the agent's task list. The item of agent $i$ is \emph{complete}, if $\textit{tasks}_{i} = \emptyset$. All agents have to coordinate to avoid conflicts to ensure fast completion of all tasks. The goal is to maximize the value of $\textit{score}_{t} = |\mathcal{D}_{\textit{complete}}| - \textit{tasks}_{t} - \textit{cost}_{t} - \textit{tpen}_{t}$, where $|\mathcal{D}_{\textit{complete}}|$ is the number of complete items, $\textit{tasks}_{t}$ is the total number of unprocessed tasks, $\textit{cost}_{t}$ is the total sum of processing cost at each machine, and $\textit{tpen}_{t}$ is the total sum of time penalties of 0.1 per incomplete item at every time step until $t$.
The global reward $r_{t} = \textit{score}_{t+1} - \textit{score}_{t} = \sum_{i = 1}^{N} r_{t,i}$ can be decomposed into local rewards $r_{t,i}= \textit{score}_{t+1,i} - \textit{score}_{t,i}$, where $\textit{score}_{t,i}$ is calculated similarly to $\textit{score}_{t}$ by only regarding the tasks, time penalties, and machine costs concerning agent $i$ itself.

\subsection{Methods}

\subsubsection{Multi-Agent Planning}\label{subsubsec:multi_agent_planning_methods}
We implemented different planning approaches to evaluate the most suited approach for training.

\paragraph{DOLUCT Variants}
We instantiated DOLUCT (Section \ref{subsec:DOLUCT}) with different value functions $\hat{V}$ and policies $\hat{\pi}_{i}$ to simulate other agents:
\begin{itemize}
\item $\textit{DOLUCT}_{\textit{Baseline}}$ uses a random uniform policy $\hat{\pi}_{i,\textit{Random}}$ and a value function $\hat{V}(s_{t}) = 0$ for each state $s_{t} \in \mathcal{S}$.
\item $\textit{DOLUCT}_{\textit{Random}}$ uses a random uniform policy $\hat{\pi}_{i,\textit{Random}}$ and approximates $V^{*}$ (Section \ref{subsubsec:marl_methods}) to evaluate leaf states \cite{silver2017mastering}.
\item $\textit{DOLUCT}_{\textit{STEP}}$ uses the current policy and value function approximation of $f_{i,\theta} = \langle \hat{\pi}_{i}, \hat{V} \rangle$ as described in Algorithm \ref{algorithm:STEP} with $\textit{DPlan} = \textit{DOLUCT}$.
\end{itemize}

\paragraph{Decentralized MCTS (DMCTS)}
We implemented $\textit{DMCTS}_{\textit{STEP}}$, as a closed-loop planner using STEP similarly to $\textit{DOLUCT}_{\textit{STEP}}$ (with $\textit{Nd}_{t} = s_{t} \in \mathcal{S}$ in Eq. \ref{eq:ucb1_STEP} and $\textit{DPlan} = \textit{DMCTS}$ in Algorithm \ref{algorithm:STEP}). Unlike $\textit{DOLUCT}_{\textit{STEP}}$, $\textit{DMCTS}_{\textit{STEP}}$ conditions its action selection on states and state transitions caused by simulated joint actions with a worst-case branching factor of $b_{\textit{cl}} = |\mathcal{S}|\cdot|\mathcal{A}_{i}|$ (Fig. \ref{fig:closed_loop_planning} and Section \ref{subsec:DOLUCT}).

\paragraph{Direct Cross Entropy (DICE) Planning}
Based on \cite{phan2018evade}, we implemented an open-loop version of $\textit{DICE}$ \cite{oliehoek2008cross} to perform centralized planning on the joint action MDP formulation of the MMDP. Our implementation approximates $V^{*}$ (Section \ref{subsubsec:marl_methods}) to evaluate leaf states \cite{silver2017mastering,phan2018evade}. Unlike DOLUCT and DMCTS, where all agents perform independent local planning in parallel, DICE is not parallelizable.

\subsubsection{Policy and Value Function Approximation}\label{subsubsec:neural_network_architecture}
We used deep neural networks $\hat{\pi}_{\phi}$, $\hat{Q}_{\omega}$, $\hat{V}_{\rho}$, and $f_{\theta}$ with weights $\phi$, $\omega$, $\rho$, and $\theta$
to implement different MARL algorithms (Section \ref{subsubsec:marl_methods}).
All networks receive the \emph{global state} $s_{t}$ and the \emph{individual state} information $s_{t}^{i}$ of agent $i$ as input, which we refer to as $s_{t,i} = \langle s_{t}, s_{t}^{i} \rangle$ to omit the index $i$ of the function approximators. An experience buffer $E$ was implemented to store the last 10,000 transitions and to sample minibatches of size 64 to perform stochastic gradient descent (SGD) using ADAM with a learning rate of 0.001. $E$ was initialized with 5,000 experience samples generated with $\textit{DOLUCT}_{\textit{Baseline}}$ ($n_{b} = 512$). We set $\gamma = 0.95$. All value-based approaches use a \emph{target network}, where a neural network copy with parameters $\omega^{-}$ or $\rho^{-}$ is used to generate the TD targets \cite{mnih2015human}. The target network is updated every 5,000 SGD steps with the trained parameters $\omega$ or $\rho$ resp.

The global and individual state can be represented as multi-channel image for both domains. The global state input $s_{t}$ is a tensor of size $8 \times 8 \times 3$ (PE) or $5 \times 5 \times 35$ (SF). The individual state input $s_{t}^{i}$ is a tensor of size $8 \times 8 \times 3$ (PE) or $5 \times 5 \times 3$ (SF). The feature planes, used for $s_{t}$ and $s_{t}^{i}$, are described in Table \ref{tab:feature_plane_stack} and \ref{tab:feature_plane_stacks} resp.

\begin{table*}
{\small
\centering
\caption{Description of all feature planes as global state input $s_{t}$, which are adopted from \cite{gupta2017cooperative} (PE) and \cite{phan2018evade} (SF) resp.}
\begin{tabular}[center]{|p{2.1cm}|p{2.30cm}|P{1.05cm}|p{10.5cm}|} \hline
Domain & Feature & \# Planes & Description \\ \hline
\multirow{3}{*}{Pursuit \& Evasion} & Pursuer position & 1 & The position of each pursuer in the grid (e.g., red circles in Fig. \ref{fig:pursuit_evasion_map}). \\
& Evader position & 1 & The position of each evader in the grid (e.g., blue circles in Fig. \ref{fig:pursuit_evasion_map}). \\
& Obstacle position & 1 & The position of each obstacle in the grid (e.g., gray rectangles in Fig. \ref{fig:pursuit_evasion_map}). \\ \hline
\multirow{4}{*}{Smart Factory} & Machine type & 1 & The type of each machine as a value between 0 and 14 (Fig. \ref{fig:machine_grid}) \\
& Agent state & 4 & The number of agents standing at machines whose types are (not) contained in their current bucket of tasks and whether they are enqueued or not. \\
& Tasks ($1^{\textit{st}}$ bucket) & 15 & Spatial distribution of agents with a matching task in their first bucket for each machine type. \\
& Tasks ($2^{\textit{nd}}$ bucket) & 15 & Same as "Tasks ($1^{\textit{st}}$ bucket)" but for the second bucket. \\ \hline
\end{tabular}\label{tab:feature_plane_stack}
}
\end{table*}

\begin{table}
{\small
\centering
\caption{Feature planes as individual state input $s_{t}^{i}$.}
\begin{tabular}[center]{|c|c|c|} \hline
Plane & Pursuit \& Evasion & Smart Factory \\ \hline
1 & Agent $i$'s position & Agent $i$'s position \\
2 & Evader positions & Machine positions ($1^{\textit{st}}$ bucket) \\
3 & Obstacle positions & Machine positions ($2^{\textit{nd}}$ bucket) \\ \hline
\end{tabular}\label{tab:feature_plane_stacks}
}
\end{table}

Both input streams are processed by separate residual towers of a convolutional layer with 128 filters of size $5 \times 5$ with stride 1, batch normalization and two subsequent residual blocks. Each residual block consists of two convolutional layers with 128 filters of size $3 \times 3$ with stride 1 and batch normalization. The input of each residual block is added to the corresponding normalized output. The concatenated output of both residual towers is processed by a fully connected layer with 256 units. All hidden layers use ReLU activation. The residual network architecture was inspired by \cite{silver2017mastering}. The units and activation of the output layer depend on the approximated function and are listed in Table \ref{tab:value_network_hyperparameter}. $f_{\theta}$ has the same outputs as $\hat{\pi}_{\phi}$ and $\hat{V}_{\rho}$ but combined into a single neural network.

\begin{table}
{\small
\centering
\caption{Output layer of the neural networks.\label{tab:value_network_hyperparameter}}
\begin{tabular}[center]{|l|l|c|l|} \hline
Neural Network & Layer Type & \# units & activation \\ \hline
$\hat{\pi}_{\phi}$ & fully connected & $|\mathcal{A}_{i}|$ & softmax\\ 
$\hat{Q}_{\omega}$ & fully connected & $|\mathcal{A}_{i}|$ & linear\\
$\hat{V}_{\rho}$ & fully connected & 1 & linear\\\cline{2-4}
$f_{\theta}$ & \multicolumn{3}{ l| }{same as $\hat{\pi}_{\phi}$ and $\hat{V}_{\rho}$ combined into one network} \\ \hline
\end{tabular}
} 
\end{table}

\subsubsection{Multi-Agent Reinforcement Learning}\label{subsubsec:marl_methods}
We implemented different MARL algorithms with deep neural networks (Section \ref{subsubsec:neural_network_architecture}). We performed \emph{centralized learning}, where a single neural network is trained for all agents, but \emph{decentralized execution}, where the trained neural network is deployed on each agent for testing \cite{foerster2016learning,gupta2017cooperative}.

\begin{itemize}
\item \emph{Strong Emergent Policy approximation (STEP)} is learned as explained in Section \ref{subsec:STEP_approach} and Algorithm \ref{algorithm:STEP}, where $f_{\theta}$ minimizes $L_{\textit{STEP}}$ according to Eq. \ref{eq:step_loss_function} for each $e_{t} \in E$.
\item \emph{Distributed V-Learning (DVL)} approximates $V^{*}$ of the optimal joint policy $\pi^{*}$ with $\hat{V}_{\rho}$  by minimizing $L_{\textit{DVL},i} = (r_{t} + \gamma \hat{V}_{\rho^{-}}(s_{t+1,i}) - \hat{V}_{\rho}(s_{t,i}))^{2}$ for each $e_{t} \in E$ \cite{sutton1988learning,sutton1998introduction,phan2018evade}.
\item \emph{Distributed Q-Learning (DQL)} approximates $Q_{i}^{*}$ with $\hat{Q}_{\omega}$ by minimizing $L_{\textit{DQL},i} = (r_{t} + \gamma \textit{max}_{a_{t+1,i}}(\hat{Q}_{\omega^{-}}(s_{t+1,i},a_{t+1,i})) - \hat{Q}_{\omega}(s_{t,i},a_{t,i}))^{2}$ for each $e_{t} \in E$ \cite{tan1993multi,watkins1992q,mnih2015human,tampuu2017multiagent,leibo2017multi}.
\item \emph{Distributed Actor-Critic (DAC)} approximates $\pi_{i}^{*}$ with $\hat{\pi}_{\phi}$ by minimizing $L_{\textit{DAC},i} = -\textit{log}(\hat{\pi}_{\phi}(a_{t,i}|s_{t,i}))(r_{t} + \gamma\hat{V}_{\rho}(s_{t+1,i}) - \hat{V}_{\rho}(s_{t,i}))$ \cite{sutton2000policy,konda2000actor,foerster2017counterfactual}, where $\hat{V}_{\rho}$ is trained with DVL \footnote{$\hat{\pi}_{\phi}$ and $\hat{V}_{\rho}$ are trained separately, because it is more stable than training $f_{\theta}$ for DAC.}.
\end{itemize}

\subsection{Results}
We tested all approaches in settings with $N = 2,4,6$ agents for PE and $N = 4,8,12$ agents for SF. An \emph{episode} is reset after 50 time steps, when all evaders are captured (PE), or when all items are complete (SF). A \emph{run} consists of 500 episodes and is repeated 30 times. As a no-learning algorithm, $\textit{DOLUCT}_{\textit{Baseline}}$ ($n_{b} = 512$) was run 100 times to determine its average performance for comparison.

The performance for PE is evaluated with the evader \emph{capture rate} $R_{\textit{capture}} = \frac{\textit{N}_{\textit{captured}}}{N}$, where $\textit{N}_{\textit{captured}}$ is the number of captured evaders and $N$ is the total number of evaders.

The performance for SF is evaluated with the value of $\textit{score}_{50}$ and the item \emph{completion rate} $R_{\textit{completion}} = \frac{|\mathcal{D}_{\textit{complete}}|}{N}$.

\begin{table*}
{\small
\centering
\caption{Average capture rate $R_{\textit{capture}}$ (Pursuit \& Evasion) and completion rate $R_{\textit{completion}}$ (Smart Factory) at the end of the $500^{th}$ episode of all experiments within a 95\% confidence interval.}
\begin{tabular}[center]{|L{3.6cm}||P{1.8cm}|P{1.8cm}|P{1.8cm}||P{1.8cm}|P{1.8cm}|P{1.8cm}|} \hline
& \multicolumn{3}{ P{5.4cm}|| }{Pursuit \& Evasion ($R_{\textit{capture}}$)} & \multicolumn{3}{ P{5.4cm}| }{Smart Factory ($R_{\textit{completion}}$)} \\ \hline
\# agents $N$ & 2 & 4 & 6 & 4 & 8 & 12\\ \hline
\# states $|\mathcal{S}|$& $\approx 4.1 \cdot 10^{6}$ & $\approx 1.7 \cdot 10^{13}$ & $\approx 6.9 \cdot 10^{19}$ & $\approx 6.9 \cdot 10^{53}$ & $\approx 4.8 \cdot 10^{107}$ & $\approx 3.3 \cdot 10^{161}$ \\
\# joint actions $|\mathcal{A}| = |\mathcal{A}_{i}|^{N}$ & $5^{2} = 25$ & $5^{4} = 625$ & $5^{6} = 15,625$ & $6^{4} = 1,296$ & $6^{8} \approx 1.7 \cdot 10^{6}$ & $6^{12} \approx 2.2 \cdot 10^{9}$ \\ \hline
$\textit{DOLUCT}_{\textit{STEP}}$ ($n_{b} = 128$) & $88.3 \pm 8.3 \%$ & \bm{$93.3 \pm 5.0 \%$} & $88.9 \pm 6.7 \%$ & \bm{$100 \pm 0.0 \%$} & \bm{$99.6 \pm 0.8 \%$} & \bm{$99.2 \pm 1.1 \%$} \\
$\textit{DOLUCT}_{\textit{Random}}$ ($n_{b} = 128$) & $23.3 \pm 10.0 \%$ & $9.2 \pm 5.8 \%$ & $22.2 \pm 5.6 \%$ & $96.7 \pm 3.3 \%$ & $92.5 \pm 2.9 \%$ & $88.3 \pm 1.9 \%$ \\
$\textit{DOLUCT}_{\textit{Baseline}}$ ($n_{b} = 512$) & $44.5 \pm 7.5 \%$ & $86.3 \pm 4.0 \%$ & \bm{$96.3 \pm 1.8 \%$} & $90.5 \pm 3.0 \%$ & $83.3 \pm 2.6 \%$ & $76.2 \pm 3.0 \%$ \\
$\textit{DMCTS}_{\textit{STEP}}$ ($n_{b} = 128$) & \bm{$90.0 \pm 10.0 \%$} & $90.0 \pm 6.7 \%$ & $91.7 \pm 5.6 \%$ & $98.3 \pm 2.5 \%$ & $95.8 \pm 2.1 \%$ & $85.9 \pm 4.0 \%$ \\
$\textit{DICE}$ ($n_{b} = 128N$) & $8.3 \pm 6.7 \%$ & $58.3 \pm 9.2 \%$ & $87.2 \pm 3.9 \%$ & $97.5 \pm 3.3 \%$ & $98.8 \pm 1.7 \%$ & $98.3 \pm 1.7 \%$ \\ \hline
$\textit{STEP}$ ($n_{b} = 256$) & \bm{$60.5 \pm 4.1 \%$} & \bm{$80.3 \pm 1.2 \%$} & $81.0 \pm 0.7 \%$ & \bm{$96.9 \pm 0.9 \%$} & \bm{$97.3 \pm 0.7 \%$} & \bm{$97.4 \pm 1.5 \%$} \\
$\textit{STEP}$ ($n_{b} = 128$) & $58.1 \pm 3.2 \%$ & $77.8 \pm 1.9 \%$ & $80.8 \pm 1.6 \%$ & $95.1 \pm 0.9 \%$ & $96.4 \pm 0.7 \%$ & $95.5 \pm 1.4 \%$ \\
$\textit{STEP}$ ($n_{b} = 64$) & $59.5 \pm 4.8 \%$ & $76.4 \pm 2.3 \%$ & $79.4 \pm 1.2 \%$ & $91.2 \pm 0.4 \%$ & $94.7 \pm 1.2 \%$ & $92.9 \pm 1.9 \%$ \\
$\textit{DQL}_{\textit{local}}$ & $6.9 \pm 1.8 \%$ & $69.4 \pm 5.8 \%$ & \bm{$91.0 \pm 1.4 \%$} & $72.8 \pm 6.5 \%$ & $72.5 \pm 3.2 \%$ & $70.0 \pm 7.6 \%$ \\
$\textit{DQL}_{\textit{global}}$ & $4.3 \pm 1.6 \%$ & $19.0 \pm 2.6 \%$ & $31.5 \pm 2.9 \%$ & $16.7 \pm 9.0 \%$ & $0.6 \pm 0.5 \%$ & $0.0 \pm 0.0 \%$ \\
$\textit{DAC}_{\textit{local}}$ & $17.2 \pm 2.1 \%$ & $63.3 \pm 1.6 \% $ & $81.9 \pm 1.2 \%$ & $47.8 \pm 20.7 \%$ & $49.0 \pm 21.1 \%$ & $47.9 \pm 24.2 \%$ \\
$\textit{DAC}_{\textit{global}}$ & $14.2 \pm 0.9 \%$ & $62.5 \pm 1.2 \%$ & $77.9 \pm 2.4 \%$ & $0.0 \pm 0.0 \%$ & $0.0 \pm 0.0 \%$ & $0.0 \pm 0.0 \%$ \\ \hline
\end{tabular}\label{tab:domain_complexity_performance}
}
\end{table*}

\subsubsection{STEP Training}\label{subsubsec:STEP_training}
We trained STEP with DOLUCT and DMCTS and compared them with other planning approaches (Section \ref{subsubsec:multi_agent_planning_methods}). For all decentralized algorithms, we set $n_{b} = 128$ and $c = 1$. DICE has a computation budget of $n_{b} = 128 N$, proportional to the number of agents $N$, and a planning horizon of $h = 4$ \footnote{We experimented with $h = 2,4,8$, but 4 seemed to be the best choice for DICE.}.

\begin{figure*}[!ht]
     \subfloat[Pursuit \& Evasion (2 agents)\label{fig:pursuit_evasion_2_agents_planning}]{%
       \includegraphics[width=0.27\textwidth]{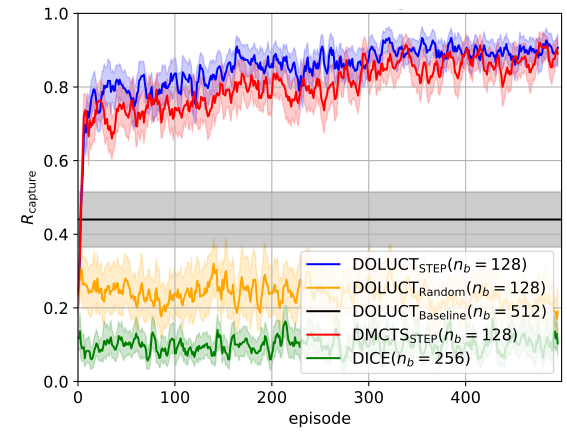}
     }
     \hfill
     \subfloat[Pursuit \& Evasion (4 agents)\label{fig:pursuit_evasion_4_agents_planning}]{%
       \includegraphics[width=0.27\textwidth]{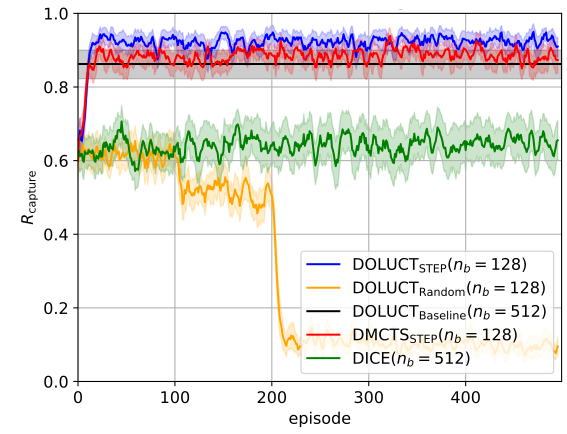}
     }
     \hfill
     \subfloat[Pursuit \& Evasion (6 agents)\label{fig:pursuit_evasion_6_agents_planning}]{%
       \includegraphics[width=0.27\textwidth]{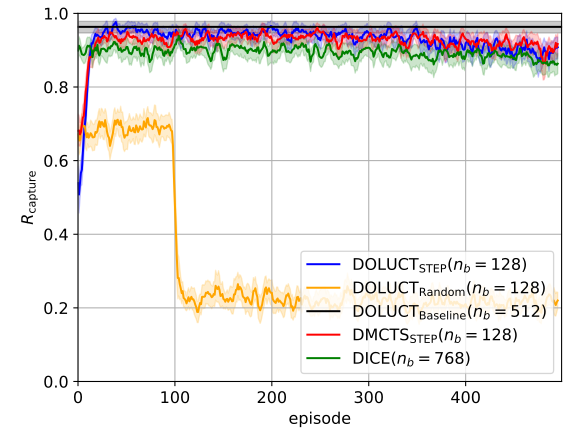}
     }
     \\
     \subfloat[Smart Factory (4 agents)\label{fig:smart_factory_4_agents_planning}]{%
       \includegraphics[width=0.27\textwidth]{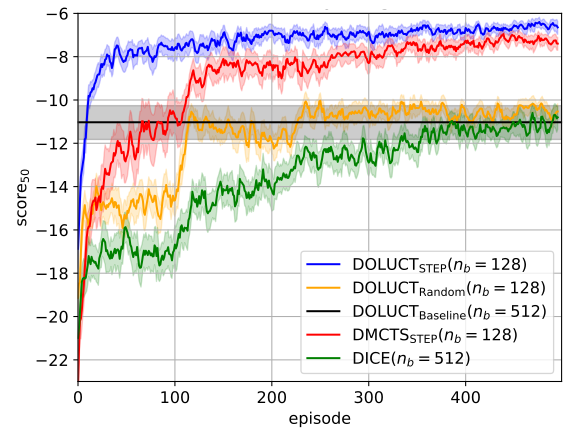}
     }
     \hfill
     \subfloat[Smart Factory (8 agents)\label{fig:smart_factory_8_agents_planning}]{%
       \includegraphics[width=0.27\textwidth]{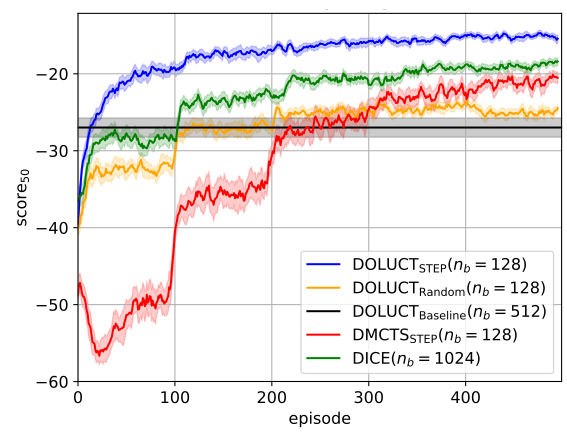}
     }
     \hfill
     \subfloat[Smart Factory (12 agents)\label{fig:smart_factory_12_agents_planning}]{%
       \includegraphics[width=0.27\textwidth]{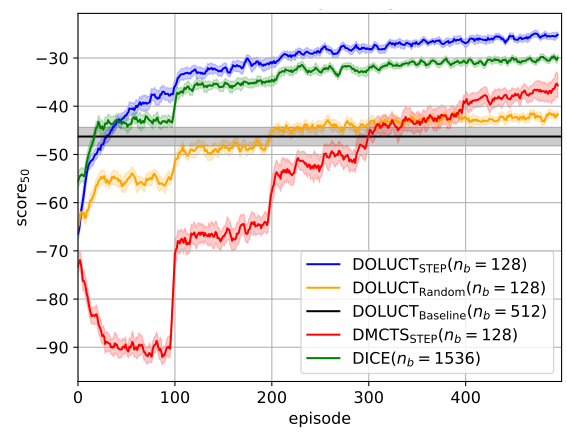}
     }
     \caption{Average training progress of $R_{\textit{capture}}$ (Pursuit \& Evasion) and $\textit{score}_{50}$ (Smart Factory) of 30 runs shown as running mean over 5 episodes for different multi-agent planning algorithms. Shaded areas show the 95 \% confidence interval.}
     \label{fig:step_planning_progress}
\end{figure*}

The results are shown in Fig. \ref{fig:step_planning_progress} and Table \ref{tab:domain_complexity_performance}. $\textit{DOLUCT}_{\textit{STEP}}$ converges fastest in all cases, except in PE with $N = 6$, and clearly outperforms all other approaches in SF. $\textit{DMCTS}_{\textit{STEP}}$ also displays progress but converges much slower than $\textit{DOLUCT}_{\textit{STEP}}$ in all SF settings. DICE only displays significant progress in SF but becomes more competitive with increasing $N$. $\textit{DOLUCT}_{\textit{Random}}$ fails to find meaningful policies in PE but slowly improves in SF.

\subsubsection{STEP Test}\label{subsubsec:STEP_test}
We evaluated the STEP approximation $f_{\theta}$ of each DOLUCT run from Section \ref{subsubsec:STEP_training} with 100 randomly generated test episodes to determine the average performance after every tenth training episode. The performance was compared with $\hat{Q}_{\omega}$ and $\hat{\pi}_{\phi}$ which were trained using DQL and DAC resp. (Section \ref{subsubsec:marl_methods}). We also implemented versions of DQL and DAC, which we call $\textit{DQL}_{\textit{local}}$ and $\textit{DAC}_{\textit{local}}$ resp., that learn with local rewards as described in Section \ref{subsec:environments} for easier multi-agent credit assignment to provide stronger baselines than $\textit{DQL}_{\textit{global}}$ and $\textit{DAC}_{\textit{global}}$ which were trained with the original global rewards. In addition, we provide the results of STEP policies that were trained with DOLUCT using computation budgets of $n_{b} = 64,256$.

\begin{figure*}[!ht]
     \subfloat[Pursuit \& Evasion (2 agents)\label{fig:pursuit_evasion_2_agents_learning}]{%
       \includegraphics[width=0.27\textwidth]{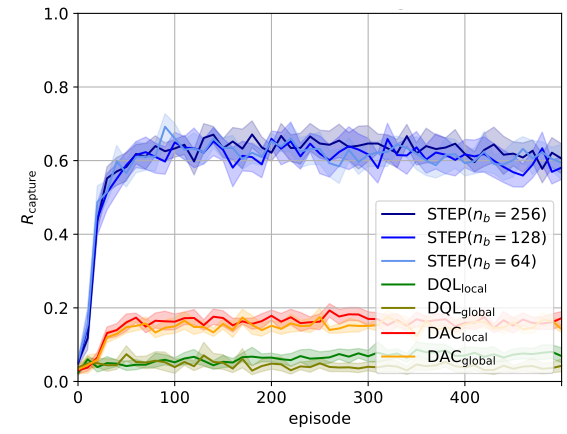}
     }
     \hfill
     \subfloat[Pursuit \& Evasion (4 agents)\label{fig:pursuit_evasion_4_agents_learning}]{%
       \includegraphics[width=0.27\textwidth]{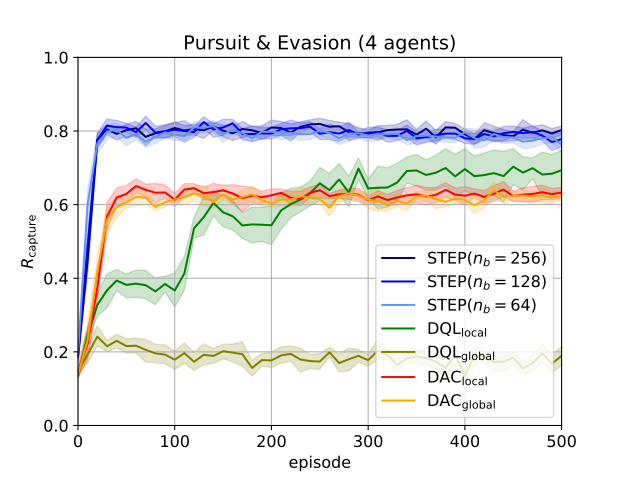}
     }
     \hfill
     \subfloat[Pursuit \& Evasion (6 agents)\label{fig:pursuit_evasion_6_agents_learning}]{%
       \includegraphics[width=0.27\textwidth]{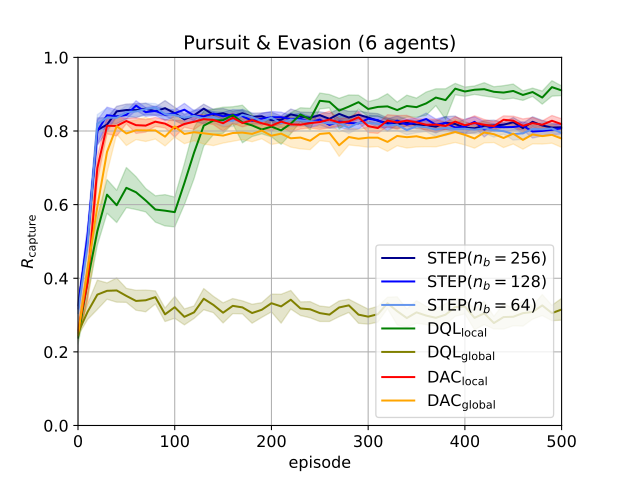}
     }
     \\
     \subfloat[Smart Factory (4 agents)\label{fig:smart_factory_4_agents_learning}]{%
       \includegraphics[width=0.27\textwidth]{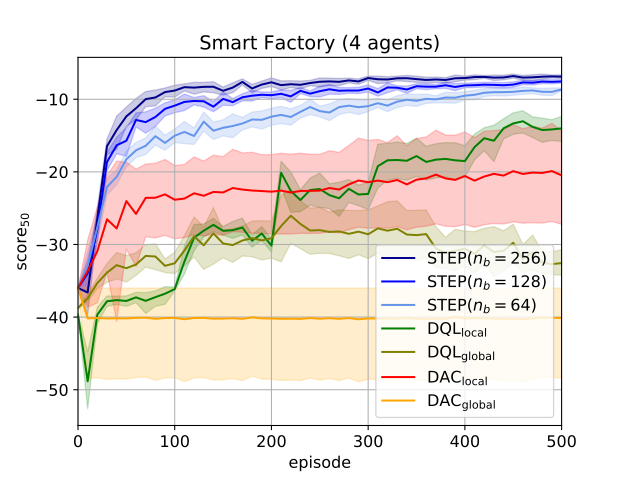}
     }
     \hfill
     \subfloat[Smart Factory (8 agents)\label{fig:smart_factory_8_agents_learning}]{%
       \includegraphics[width=0.27\textwidth]{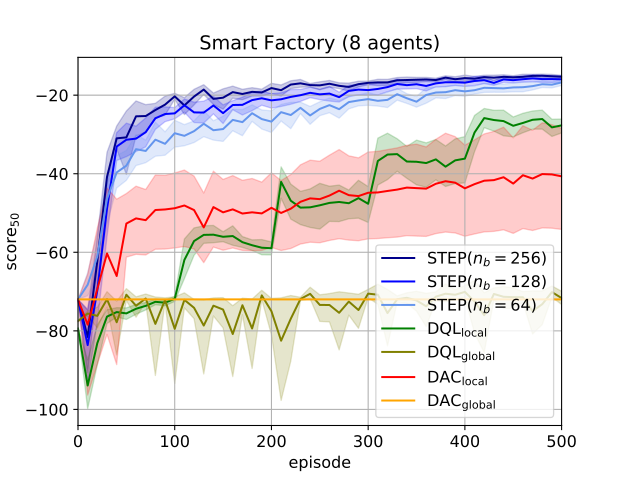}
     }
     \hfill
     \subfloat[Smart Factory (12 agents)\label{fig:smart_factory_12_agents_learning}]{%
       \includegraphics[width=0.27\textwidth]{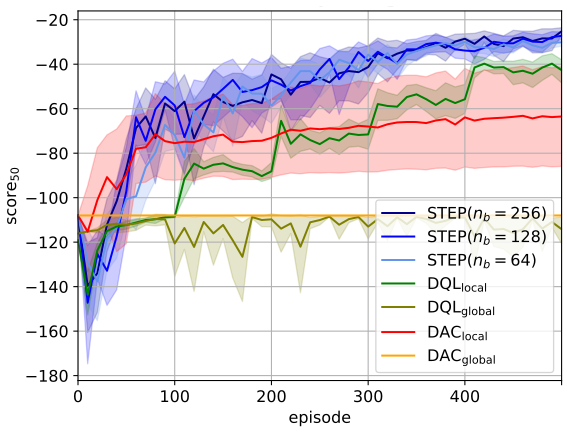}
     }
     \caption{Average test progress of $R_{\textit{capture}}$ (Pursuit \& Evasion) and $\textit{score}_{50}$ (Smart Factory) of 30 runs of the neural networks $f_{\theta}$, $\hat{Q}_{\omega}$, and $\hat{\pi}_{\phi}$ trained with STEP (using DOLUCT), DQL, and DAC resp. Shaded areas show the 95 \% confidence interval.}
     \label{fig:step_learning_progress}
\end{figure*}

The results are shown in Fig. \ref{fig:step_learning_progress} and Table \ref{tab:domain_complexity_performance}. STEP learns the strongest policies in all settings, except in PE with $N = 6$. $\textit{DQL}_{\textit{global}}$ and $\textit{DAC}_{\textit{global}}$ are unable to learn meaningful policies in SF, but $\textit{DAC}_{\textit{global}}$ shows progress in PE with increasing $N$. $\textit{DQL}_{\textit{local}}$ and $\textit{DAC}_{\textit{local}}$ always outperform their global reward optimizing counterparts but are inferior to STEP, except in PE with $N = 6$.

\subsection{Discussion}
Our experiments show the effectiveness of STEP in two domains, which are challenging in different aspects. PE has a \emph{sparse reward} structure and becomes \emph{less challenging} when the number of pursuers is large because the pursuers can distribute more effectively across the map to capture evaders. This can be seen in Fig. \ref{fig:step_planning_progress}a-c and Table \ref{tab:domain_complexity_performance}, where $\textit{DOLUCT}_{\textit{Baseline}}$ performs better with more agents.

In contrast, SF has a \emph{dense reward} structure and becomes \emph{more challenging} when the number of agents is large due to more potential conflicts at simultaneously required machines as indicated by the performance of $\textit{DOLUCT}_{\textit{Baseline}}$ in Table \ref{tab:domain_complexity_performance}, where it becomes increasingly difficult to find coordinated local policies due to the enormous search space. In PE, two pursuers need to occupy the same cell to capture an evader as a \emph{joint task}, while in SF multiple agents should \emph{avoid conflicts} by not enqueuing at the same machine.

Results from Section \ref{subsubsec:STEP_training} show that DOLUCT and DMCTS are able to improve with STEP, but DOLUCT is more suited when the problem is too large to provide sufficient computation budget (Section \ref{subsec:DOLUCT}), offering scalable performance in all domains as shown in Fig. \ref{fig:step_planning_progress} and Table \ref{tab:domain_complexity_performance}. $\textit{DOLUCT}_{\textit{STEP}}$ also outperforms the centralized DICE, which directly optimizes the joint policy but requires more total computation budget to be competitive against DOLUCT.

Results from Section \ref{subsubsec:STEP_test} show that STEP is able to learn strong decentralized policies, which can be reintegrated into the planning process to further improve and coordinate decentralized multi-agent planning in contrast to planning with just a random policy (Fig. \ref{fig:step_planning_progress} and Table \ref{tab:domain_complexity_performance}). In fact, $\textit{DOLUCT}_{\textit{Random}}$ performs very poorly in PE, where it is important to accurately predict other agents' actions to coordinate on the joint task of capturing evaders (Section \ref{subsec:scalable_policy_iteration}).

Increasing the computation budget $n_{b}$ tends to slightly improve the quality of STEP, which might be due to the decreasing bias when planning with the currently learned $f_{\theta}$ as explained in Section \ref{subsec:bias_regulation}.

In SF (Fig. \ref{fig:step_learning_progress}d-f), $\textit{DQL}_{\textit{global}}$ and $\textit{DAC}_{\textit{global}}$ are unable to adapt adequately due to much noise in the gradients caused by the global reward density. $\textit{DAC}_{\textit{global}}$ displays exploding gradients here, which leads to premature convergence towards a poor policy.
However, $\textit{DAC}_{\textit{global}}$ performs better with increasing $N$ in PE. This might be due to the sparse rewards which make the updates less noisy and the decreasing difficulty of capturing evaders, when $N$ is large. $\textit{DQL}_{\textit{local}}$ and $\textit{DAC}_{\textit{local}}$ always outperform their global reward optimizing counterparts, since their individual objectives are easier to optimize. Still, they are generally inferior to STEP because they are unable to consider emergent dependencies in the future like strategic positioning in PE or potential conflicts in SF. Unlike, these approaches, STEP is trained with decentralized multi-agent planning, which explicitly reasons about these dependencies.

\section{Conclusion and Future Work}\label{sec:conclusion}
In this paper, we proposed STEP, a scalable approach to learn strong decentralized policies for cooperative MAS with a distributed variant of policy iteration. For that, we use function approximation to learn from action recommendations of a decentralized multi-agent planner. STEP combines decentralized multi-agent planning with centralized learning, where each agent is able to explicitly reason about emergent dependencies to make coordinated decisions, only requiring a generative model for distributed black box optimization.

We experimentally evaluated STEP in two challenging and stochastic domains with large state and joint action spaces. We demonstrated that multi-agent open-loop planning is especially suited for efficient training, when the problem is too large to provide sufficient computation budget for planning. Our experiments show that STEP is able to learn stronger decentralized policies than standard MARL algorithms, without any domain or reward decomposition. The policies learned with STEP are able to effectively coordinate on joint tasks and to avoid conflicts, thus can be reintegrated into the multi-agent planning process to further improve performance.

For the future, we plan to address partially observable domains by combining multi-agent planning with deep recurrent reinforcement learning for cooperative MAS \cite{emery2004approximate,amato2015scalable,foerster2016learning,gupta2017cooperative,foerster2017counterfactual}.


\bibliographystyle{ACM-Reference-Format}  
\bibliography{references}  

\end{document}